# Word Level Font-to-Font Image Translation using Convolutional Recurrent Generative Adversarial Networks

[a]Ankan Kumar Bhunia, [b]Ayan Kumar Bhunia, [c]Prithaj Banerjee, [b]Aishik Konwer, [b]Abir Bhowmick, [d]Partha Pratim Roy[*],[e]Umapada Pal

[a]Department of EE, Jadavpur University, Kolkata, India

[b]Department of ECE, Institute of Engineering & Management, Kolkata, India.

[c]Department of CSE, Institute of Engineering & Management, Kolkata

[d]Department of CSE, Indian Institute of Technology Roorkee, India

[e]CVPR Unit, Indian Statistical Institute, Kolkata, India

*email: 2partharoy@gmail.com

***Abstract*— Conversion of one font to another font is very useful in real life applications. In this paper, we propose a Convolutional Recurrent Generative model to solve the word level font transfer problem. Our network is able to convert the font style of any printed text images from its current font to the required font. The network is trained end-to-end for the complete word images. Thus it eliminates the necessary pre-processing steps, like character segmentations. We extend our model to conditional setting that helps to learn one-to-many mapping function. We employ a novel convolutional recurrent model architecture in the Generator that efficiently deals with the word images of arbitrary width. It also helps to maintain the consistency of the final images after concatenating the generated image patches of target font. Besides, the Generator and the Discriminator network, we employ a Classification network to classify the generated word images of converted font style to their subsequent font categories. Most of the earlier works related to image translation are performed on square images. Our proposed architecture is the first of its kind which can handle images of varying widths. Word images generally have varying width depending on the number of characters present. Hence, we test our model on a synthetically generated font dataset. We compare our method with some of the state-of-the-art methods for image translation. The superior performance of our network on the same dataset proves the ability of our model to learn the font distributions.**

## I. Introduction

With tremendous advancement in technology and multimedia, images and videos become sole part of our day to day life activities. So, these available images and videos become a subject of study. Image processing, computer vision and computer graphics areas deal with study of images and videos. Various works have been done in the field of image retrieval, image classification but comparatively very few works have been done in order to perform image to image translation. Isola et al. [1] proposed a novel method of such image to image translation using Conditional Generative Adversarial Networks (cGANs). In our proposed method, we have extended the task of image to image translation in Document Image Analysis (DIA) domain. Our work focuses on font to font translation of document images of printed words. To the best of our knowledge, no other previous work tried to device any method for word level font to font translation. Using our method, images of printed words written in one particular font, can be easily transformed to any other commonly used font. This makes it very useful for editing purpose with no need of soft copy. One can edit by directly taking the photograph and changing the font without re-editing them, thus saving sufficient time and effort. Sometimes fonts of old books or manuscripts become fainted, which make them difficult to understand. These fonts can be given a new fresh look, easing readability. Font-to-font translation can also be applied to graphic designs. Other useful aspects of this novel approach include designing of cover pages of magazines or books, with the advantage that different fonts can be tried without having several softcopies of the background. Fig.1 illustrates the font-to-font translation problem.

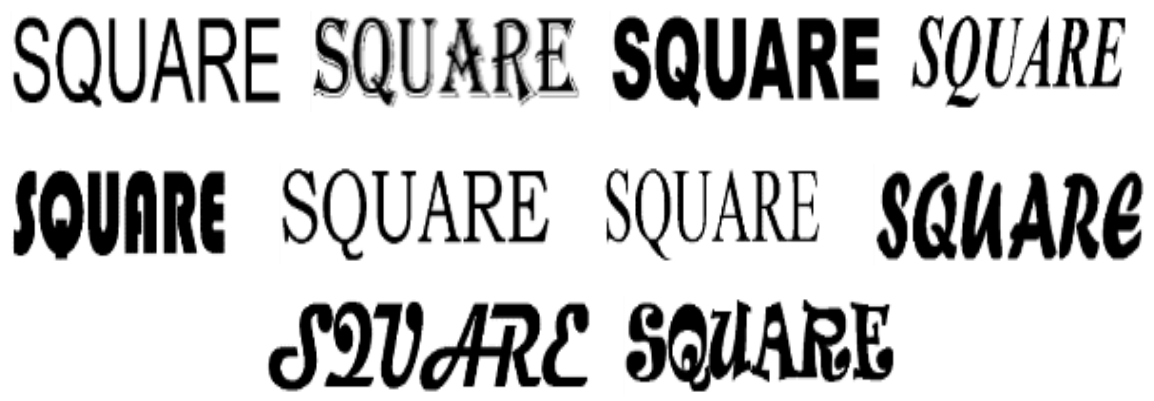

**Fig.1. Example showing the font-to-font translation**

In this field of computer vision and image processing, special-purpose machineries [2], [3] were used to deal with these kind of approaches where an image is given as input to generate corresponding images as output. Gradually with the advent of Deep Neural Network, more accuracy and perfectness could be achieved easily in these works.

In our proposed method we have used the concept of Generative adversarial networks (GANs)[4] where two neural nets are used, one to generate the images and another to discriminate between real and fake. These two neural nets in a single frame-work contest among themselves to generate more realistic fake images and thus improving the efficiency. In this work GANs are explored in conditional setting. Conditional GANs (cGANs)[5] learn a conditional generative model of data in a similar manner as normal GANs learn generative model. Thus cGANs are more appropriate for our font-to-font translation where we assign condition on an input printed font image to generate a different font image as output. GAN is a very recent concept and within this short span it is extensively used in various applications. But none of the previous papers used it for word level font-to-font translation where images can be of varying width. Our main objective in this paper is to show how this problem can be handle using conditional GANs and to device a new GANs architecture for this purpose. The use of our work is manifold ranging from day to day uses to uses in designing.

The contributions of the paper are as follows: Firstly, we develop a novel Convolutional Recurrent Generative Adversarial Network Architecture for Image to Image translation problem for the images with varying width. Secondly, we show the potential of our architecture in word level Font-to-Font translation problem and have achieved superior performance compared to existing methods in case of images with varying widths.

The rest of the paper is organized as follows. In Section II we discuss some recently developed related works for image to image translation and Generative Adversarial Networks. In Section III, the Convolutional Recurrent architecture for Generative Adversarial Networks is detailed. Section IV discusses our dataset preparation method and implementation details along with comparative study with some baseline methods for font-to-font translation task. Finally, conclusion is given in Section V.

## II Related Work

To the best of our knowledge, our method based on Generative Adversarial Networks is a novel approach in the sense that no other previous methods tried to translate images of one font to any other fonts. Few works have been done in the field of image to image translation[1], [6]. Regression or classification [7]–[9] of each pixel were used for solving image to image translation problems. Here the output space is treated as unstructured where given the input image, every output pixel is treated conditionally independent from the rest. A structured loss is learnt by cGANs instead. Joint configuration of the output is penalized by structured losses.

GANs are now used in image editing [10], image generation [11] as well as in representation learning [12]. Main reason for the success of GANs is the concept of adversarial loss which generates indistinguishable fake images. This is mainly advantageous in generation of image related applications because this is the optimization objective in computer graphics. Here we design the adversarial loss for learning the mapping in order to generate the translated font images that cannot be distinguished from target domain images given sufficient training datasets. Reed et al. proposed a novel approach of text to image synthesis using GANs[13]. Though it is tough to achieve the goal of generating realistic images from text document by current AI set ups but with the advent of strong recurrent neural nets and generic architectures, representation of discriminative text features could be learnt easily. Moreover, images of specific categories like room interiors, faces or album covers could be generated using GANs and Reed et al. in this work tried to bridge those advances in image, text translation by developing a Generative Adversarial and deep architecture by character to pixel transformation. Various other applications of GANs in the field of image processing include increasing of image resolution. Ledig et al. in their paper[14] used SRGAN i.e., a Generative Adversarial Network (GAN) for image Super-Resolution (SR) which is the first framework with the ability to infer natural images which are photo-realistic for an upscaling factor of 4x. Nguyen et al. [15] performed the gradient ascent to enhance the activations of neurons in a completely separate classifier of a generator network latent space and established an interesting path to generate unique images. In [16] a method called "Plug and Play Generative Networks" is proposed which is an extension of the previous method. This is done by the introduction of an additional prior on the latent code, which improved sample diversity as well as sample quality, producing a state-of-the-art model which generates images of higher resolution (227x227) compared to previous generative models. Shrivastava et al. [17] in their work proposed a Simulated+Unsupervised (S+U) learning model based on Adversarial Network principles, where the authenticity of a simulator's output is modified by using unlabelled real data keeping the information based on annotations from the stimulator preserved.

Previously, many works have been done in conditional GANs. Major and most recent works which include conditioned GANs are on images, text[13] and discrete labels . Generation of images from sparse annotations [18], [19],future frame prediction [20], prediction of images from normal map [21],product photo generation [22] are some of the most current applications of image conditional models. Various other works have used GANs unconditionally for image to image translation. All of these papers have gained fruitful results on future state prediction [23], style transfer [24] ,user constraints guided manipulation of image [10]and super resolution [14]. These methods have certain specific applications.

Some works used a U-Net based architecture [25] where as some used a convolution PatchGAN classifier as the discriminator to penalize structures at image patches scale. For capturing local style statistics one such PatchGAN architecture was devised in [24]. These works also showed the effectiveness of their approaches on wider range of problems by considering the effect of changing patch size. Numerous applications of GANs are done in various problems but our method is unique as it is concerned to font to font translation where word images can be of varying width. Moreover, our framework is simple and the same architecture can be extended for image to image translation task where width of the images varies extensively.

## III PROPOSED FRAMEWORK

In this section, we present our proposed fontGAN network. A brief overview of the proposed model is illustrated in Fig.2. The model is composed by three main parts. These parts are: (1) A generator network that transfers a text image from one font style to another font style conditioned on respective font categories, (2) the discriminator network which tries to distinguish the real/fake font images, and (3) the classification network that classifies the generated font to the subsequence categories. In subsection A, we first briefly review the Generative adversarial networks (GAN). In subsection B, we describe how a GAN model can be used in the font-to-font translation problems.

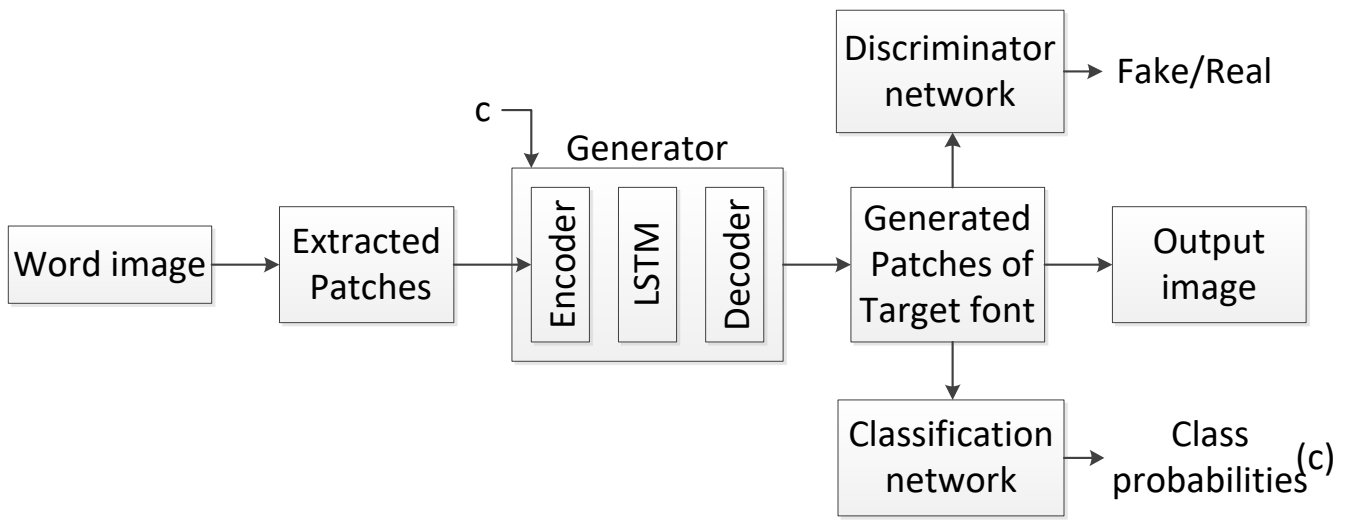

**Fig.2. Flowchart of our framework**

### A. *Generative Adverserial Networks*

A GAN[4] framework is composed of two models, the Generative model G and the discriminative model D, both are parameterized by neural networks. The generator aims to learn a mapping function from a prior noise distribution $p_z$ to an unknown data distribution $p_{data}$ in the real data space. The discriminator tries to distinguish between real and generated data. Both networks are simultaneously trained competing against each other in a min-max game with value function $V(G,D)$:

$$\min_G \max_D V(G,D) = \mathbb{E}_{x\sim p_{data}}[\log(\mathrm{D}(\mathrm{x})] + \mathbb{E}_{z\sim p_z}\left[\log\left(1-\mathrm{D}\left(\mathrm{G}(\mathrm{z})\right)\right)\right] \quad \ldots\ldots\ldots (1)$$

During training, the generator learns to generate more realistic images to fool the discriminator while the discriminator improves to distinguish the real images from the generated one. This GAN model is mainly focused on learning one-to-one mappings from input to output. Conditional GANs [26] extends the vanilla GAN to a probabilistic one-to-many mapping method where the discriminator and the generator are conditioned on some extra auxiliary information c (i.e. class labels). The objective function for conditional GAN would be:

$$\min_G \max_D V(G,D) = \mathbb{E}_{x\sim p_{data}}[\log(\mathrm{D}(\mathrm{x}|\mathrm{c})] + \mathbb{E}_{z\sim p_z}\left[\log\left(1-\mathrm{D}\left(\mathrm{G}(\mathrm{z}|\mathrm{c})\right)\right)\right] \quad \ldots\ldots\ldots (2)$$

### B. *Font Transfer GANs*

In this section, we describe our word level end-to-end font transfer framework. Generating images containing text is a challenging problem where consistency of the text regions is a major objective for better readability. In our case, as the size of the input word image is variable it becomes difficult to train the traditional GAN model. We propose a Recurrent GAN architecture which is specially designed to deal with images of arbitrary width. It is trained in Conditional setting that will help to extend our model to one-to-many mapping method.

The objective of the font transfer problem is that the model takes word image of specific font type as input and returns the word image in a different font style. At first, the text images are resized to a fixed height of 32 pixels keeping the aspect ratio same. Then patches of size 32×32 are extracted from the images. We represent each text image as a sequence of extracted patches: $X_i = \{x_{i,1}, x_{i,2}, x_{i,3}, \ldots, x_{i,N_i}\}$. The patches are then fed to a Generator network $G$.

The generator consists of an encoder, a LSTM and a decoder network. The Encoder network encodes each 32×32 image patch $x$ to a latent vector $z$ of dimension 256. Thus we obtain the feature representation for all the patches: $Z_i = \{z_{i,1}, z_{i,2}, z_{i,3}, \ldots, z_{i,N_i}\}$. The sequence of features is fed to a Recurrent Neural Network model (RNN). RNN is able to handle a sequence of arbitrary length. A basic RNN unit takes a frame of input sequence $z_t$ and its initial hidden state $h_{t-1}$ as input at a particular time-step $t$ and updates its hidden sate value $h_t$ with a non-linear function: $h_t = g(z_t, h_{t-1})$. The basic RNN suffers from the vanishing gradient problem [27]. To learn longer dependencies we use the Long short-term memory (LSTM) network [28]. The internal gate structure of the LSTM network helps to overcome the vanishing gradient problems. We adopt bidirectional LSTM [29] to learn the dependencies in both directions. The LSTM network returns a feature sequence $H_i = \{h_{i,1}, h_{i,2}, h_{i,3}, \ldots, h_{i,N_i}\}$. Each feature vector in the feature sequence actually corresponds to a patch in the original word image. A decoder network is used to generate word image patches of target font style from the feature vectors. The final output image can be obtained by concatenating the patches along the width. Although, this method is efficient at converting word images from one font to another font, but may fail to maintain the consistency along the joints in the concatenated output image. To overcome this problem, the output feature vector from the previous cell is concatenated with the input at current time step in the LSTM unit of the Generator network. It is a common practice in language models [30] to feed the output from the previous cell. It helps to maintain the consistency between the sequentially generated two successive image patches.

Hence, we use the Generator network to learn a mapping function from real samples $X$ to generated samples $Y_{gen}$. $Y_{gt}$ represents corresponding ground truth. The discriminator network $D$ is a CNN network that is used to evaluate how good the Generator network is in generating fake samples. The Discriminator inputs all the generated patches and tries to distinguish between the real and generated patches.

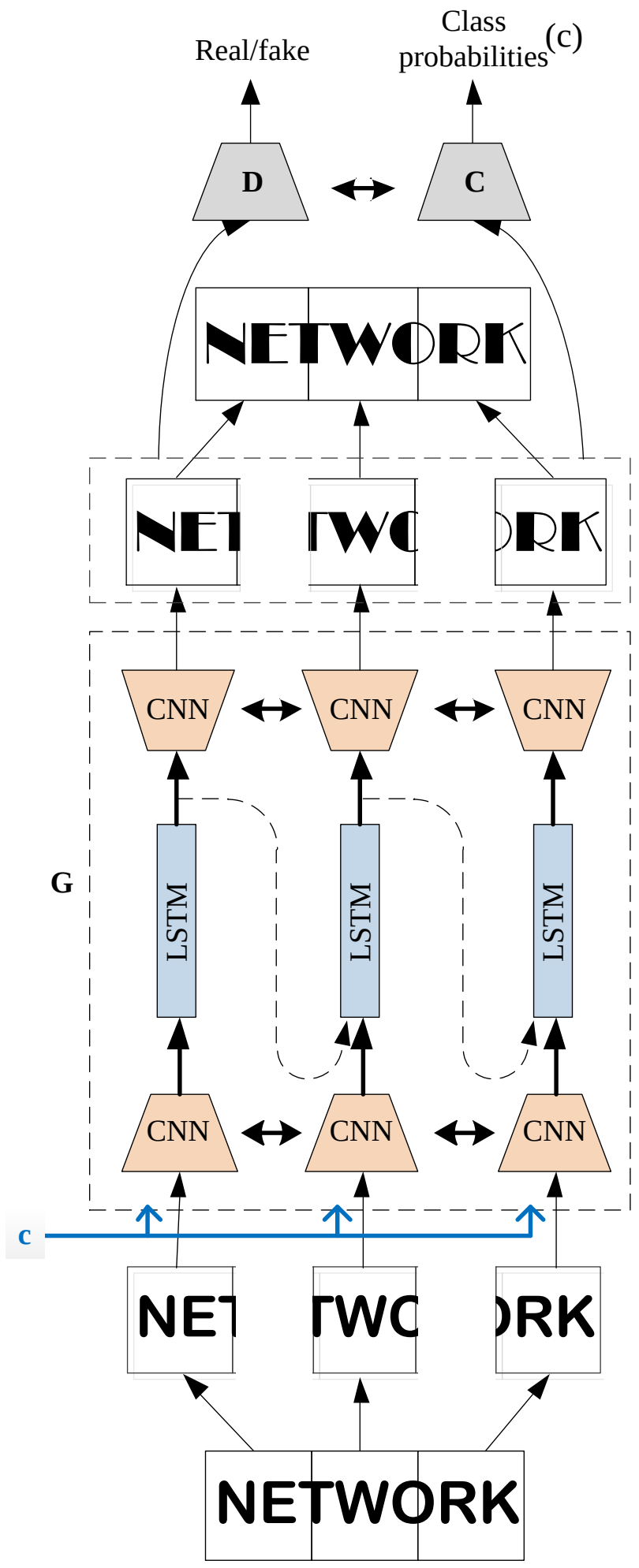

**Fig.3. Detailed architecture of our proposed framework**

A classification network $C$ is employed to measure the class probability of each patch image. The network C classifies each patch to their subsequent font category. The output of the network is a n dimensional vector, where n is the number of font classes. A Softmax function is required to obtain the class probabilities $P(c|x)$ from the output. The parameters of the classification network and the discriminator network are shared except the last two fully connected layers. Fig.3 illustrates the detailed architecture of our proposed framework.

The generator $G$ and the discriminator $D$ compete in a min-max game. D tries to minimize the following loss function for $j^{th}$ patch image:

$$L_D^{(j)} = -\mathbb{E}_y[\log(\mathrm{D}(y_j)] - \mathbb{E}_x\left[\log\left(1 - \mathrm{D}\left(\mathrm{G}(x_j)\right)\right)\right] \quad \ldots\ldots\ldots(3)$$

Whereas, G is trained to minimize $L_G^{(j)}$:

$$L_G^{(j)} = -\mathbb{E}_x\left[\log\left(\mathrm{D}\left(\mathrm{G}(x_j)\right)\right)\right] \quad \ldots\ldots\ldots(4)$$

And the classification network $C$ tries to minimize the cross-entropy loss:

$$L_C^{(j)} = -\mathbb{E}_x\left[\log\left(P(c|x_j)\right)\right] \quad \ldots\ldots\ldots(5)$$

An additional L1 loss function is used to force the model to generate images that are similar to the target images. We compute this loss by taking L1 distance between the final generated word image $Y_{gen}$ and the ground truth image $Y_{gt}$. In the literature[1], it is shown that it is beneficial to add the L1 distance loss function.

$$L_{L1} = -\mathbb{E}_{x,y}\left[\left\|Y_{gt} - Y_{gen}\right\|_1\right] \quad \ldots\ldots\ldots(6)$$

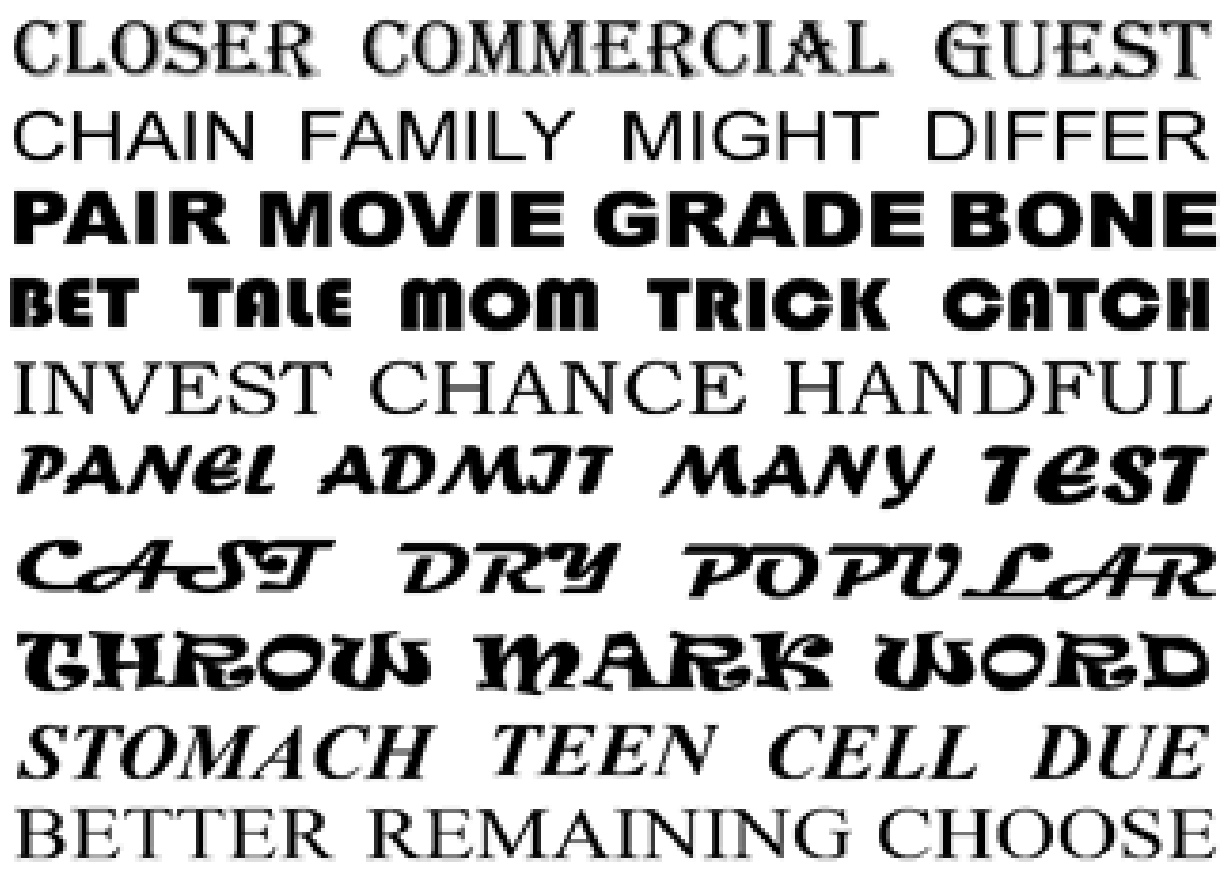

**Fig.4. Some examples of the dataset. The dataset contains fine-grained word images of varied length and various font styles - Algerian, Arial, Arial Black, Bauhaus, Bookman old style, Forte, Magneto, Ravie, Times new Roman, Times Black (top to bottom). The examples of each font are shown above.**

## IV. EXPERIMENT AND RESULT ANALYSIS

### *A. Datasets*

There is no publicly available Font dataset to evaluate our model. We have synthetically created a gray-scale word level Font dataset of 10 types of different font categories. 10,000 unique English words are processed to create word images of 10 different font styles. So, the dataset contains total 10x10,000 word images. The dataset is divided into training set (60%), validation set (20%) and test set (20%). The images are pre-processed by cropping the text region and maintaining the height of the words same for all the images. A few examples of the images are shown in Fig.4.

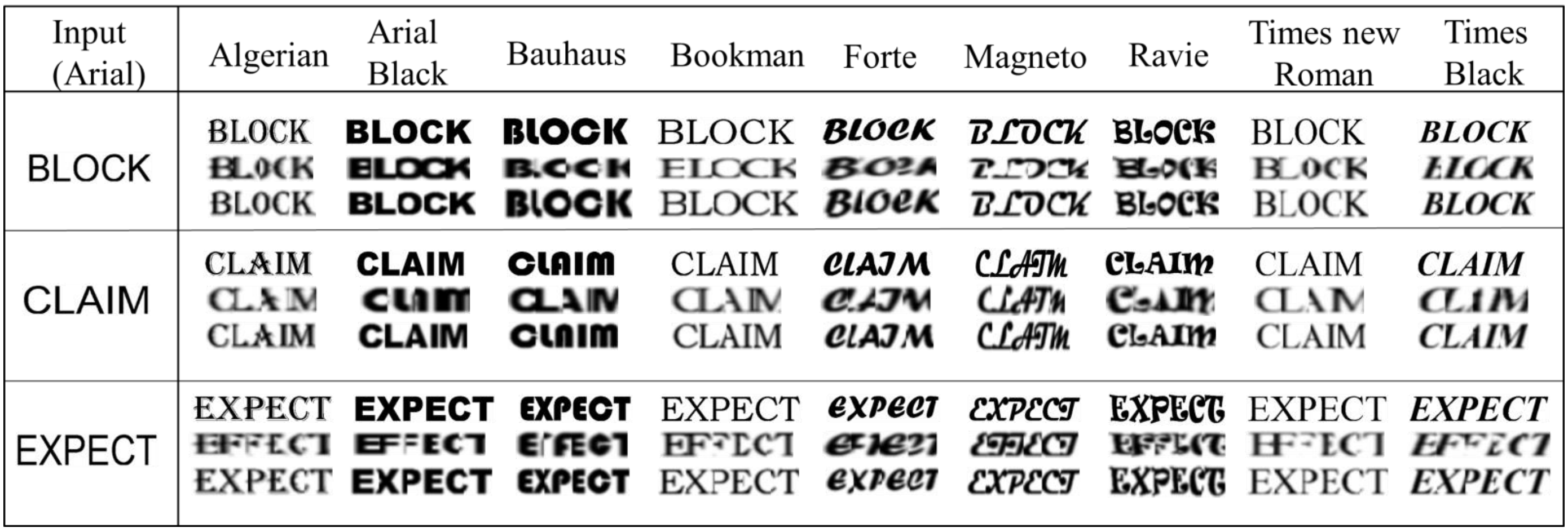

**Fig.5. Illustrations of our experimental results. The input images of Arial font and the corresponding generated output of different font styles using our model and baseline approach are shown above. 1st row contains the ground truth of different font images, the generated images using baseline method and our method are shown in 2nd row and 3rd row respectively. Please see the PDF-version of this paper to clearly understand the difference between the images generated from baseline method and our proposed method.**

*B. Model Architectures*

The Generator network is composed of an encoder network, a recurrent network and a Decoder network. The encoder inputs an image of size 32×32 and outputs a latent vector of dimension 256. The encoder network is composed of 5 convolutional layers with kernel size 5x5 and stride 2x2. The recurrent network is a LSTM network. Two bidirectional LSTMs of 512 hidden units are stacked to have higher abstraction ability. The LSTM network takes the sequence of latent vector and outputs another sequence of latent vector of dimension 256. Next, the Decoder decodes the latent vector to their corresponding image patches of size 32×32. The decoder network consists of 5 de-convolutional layers. A final tanh layer outputs the target image patch. Lastly, the discriminator network and the classification network inputs the generated image patches of size 32×32. These two networks have same configuration and the parameters are shared except the last two fully connected layers. We used leaky ReLU instead of normal ReLU in the network architectures. We applied batch normalization after each layer in the Encoder and the Decoder network.

*C. Baseline Methods*

Our baseline is a simple patch level image translation model inspired from pix2pix architecture [1]. In the baseline the generator is composed by encoder and decoder network. We will not consider the intermediate recurrent model for the baseline. The encoder network encodes each patch image to a latent representation and the decoder network decodes it to the target image patch. Comparing with this framework we will see how the proposed recurrent GAN model improves the translation performance.

*D. Training Details:*

We have done the experiments by considering Arial font as the source font of the word images. The ground truth word images of the target font styles are rescaled to a same dimension as the source image. Experiments are conducted on a server with Nvidia Titan X GPU with 12 GB of memory. We implemented the model using TensorFlow. Back propagation through time (BPTT) and Adam optimizer with learning rate 0.001 is used to optimize the objective function. The model is trained for 60 epochs with batch size 32. It takes around 2-3 hours for training in the Nvidia Titan X GPU system. L2 regularization is applied to the weights of the networks. The words images with similar aspect ratio are integrated into one batch. It will help the model to converge fast. During training, it is noticed that sometimes the discriminator becomes too strong and overpowers the generator resulting an adversarial mode collapse. Thus the training of the generator will be unstable. To overcome this problem, we pre-trained the generator with only L1 distance loss for 2-3 epochs. Then the whole model is trained end-to-end.

*E. Results*

The font translation results of our proposed model are shown in Fig.5. We compare our results with the baseline approach. The word images are chosen randomly from our test dataset. Our model learns to capture the font distributions. Some fonts are difficult to learn. But our model is strong enough to convert the font styles. In the baseline approach the font styles of the word image patches are independently converted to their target font-style. Thus, the final concatenated images suffer from the consistency problem. But in our case, the intermediate recurrent model of the network correlates the sequentially generated image patches and maintains the consistency. Feeding the previous output feature vector with the input at current time-step in the LSTM unit of the generator also helps to improve the result.

## V. Conclusions

In this paper, we proposed a Font transfer architecture based on Convolutional Recurrent Generative adversarial network. We synthetically generated the font dataset to test our model. The superior performance of our framework on that dataset demonstrates the ability in the font translation problem. Our model is able to transfer fonts at word level. It eliminates the necessary pre-processing steps such as character segmentations. Our model is efficient in converting a word image of a particular source font style. But, for a different source font the model needs to be pre-trained. It is the main limitation of our framework. In future we will extend our work by considering a general font transfer model where conversion of any source font-style could be possible. Our model is script dependent. A model trained on the English dataset will not work on a dataset of different language. We will also try to model a script independent font transfer framework in future.